\documentclass[11pt]{article}

\usepackage[preprint]{acl}

\usepackage{times}
\usepackage{latexsym}

\usepackage[T1]{fontenc}

\usepackage[utf8]{inputenc}

\usepackage{microtype}

\usepackage{inconsolata}

\usepackage[normalem]{ulem}
\usepackage{xcolor}
\usepackage{graphicx}
\usepackage{amsmath}
\usepackage{booktabs}
\usepackage{amssymb}

\usepackage{fvextra}

\DefineVerbatimEnvironment{Prompt}{Verbatim}{
  breaklines=true,
  breakanywhere=true,
  fontsize=\scriptsize,
  baselinestretch=0.95,
  breaksymbolleft={},
  breakindent=1em
}

\title{Context-CoT: Enhancing Context Learning via \\ High-Quality Reasoning Synthesis}

\author{
 \textbf{Hongbo Jin$^{1}$\thanks{Equal contribution}} \quad
 \textbf{Mingnan Zhu$^{2}$\footnotemark[1]} \quad
  \textbf{Jingqi Tian$^{3}$} \quad
  \textbf{Xu Jiang$^{1}$} \quad \\
  \textbf{Zhongjing Du$^{1}$} \quad
    \textbf{Haoran Tang$^{1}$} \quad
    \textbf{Siyi Xie$^{1}$} \quad
  \textbf{Qiaoman Zhang$^{1}$} \quad
 \textbf{Jiayu Ding$^{1}$\thanks{Corresponding author}}
\\
 $^1$Peking University,
 $^2$Xiamen University,
 $^3$Tsinghua University
}

\begin{document}
\maketitle
\begin{abstract}
While LLMs excel at reasoning over prompts using static pre-trained knowledge, they struggle significantly with context learning—the ability to dynamically extract, internalize, and apply new knowledge from complex, task-specific contexts. Recent evaluations on the CL-Bench reveal a critical capability gap: frontier models solve only 17.2\% of context-dependent tasks on average.
To bridge this gap, we propose \textbf{Context-CoT},
a novel Chain-of-Thought (CoT) data synthesis and fine-tuning framework specifically designed to enhance context learning in open-source LLMs. We construct a high-quality, context-grounded CoT dataset using a newly introduced three-stage pipeline: (i) multi-stage CoT sampling, which first guides the model to distill the long context into task-relevant intermediate representations before reasoning over the extracted contextual evidence; (ii) rubric-based minimum-leakage filtering, which hides reference answers and full rubrics during CoT generation, provides only minimal failed-rubric feedback when necessary, and filters out trajectories that violate context-specific criteria; and (iii) student-aware CoT selection, which ensures the retained CoT paths align naturally with the target model's distribution for optimal learning efficiency. Extensive experiments demonstrate that open-source models fine-tuned on our dataset achieve significant performance gains on CL-Bench, substantially reducing context-neglect errors. Our work provides a scalable, data-driven solution to transition open-source models from simple prompt followers to robust context learners. 
\end{abstract}

\section{Introduction}
\label{sec:intro}
Large Language Models (LLMs) have achieved remarkable success in tasks that require reasoning over static, pre-trained knowledge~\cite{brown2020gpt3,guo2025deepseek}.
Driven by techniques such as In-Context Learning (ICL) and instruction tuning, modern LLMs can seamlessly follow prompts and solve complex mathematical, coding, and logical problems~\cite{dong2024surveyicl,ouyang2022training,wei2022chain,chen2021evaluating, li2025system2}.
However, real-world deployment frequently presents scenarios that extend far beyond the scope of models' pre-trained corpora~\cite{dou2026clbenchlife}.
In domains such as legal adjudication,
industrial troubleshooting, or scientific discovery,
models must dynamically acquire and apply knowledge from complex provided contexts.
This fundamental capability—learning genuinely new information from a specific context and reasoning over it, rather than relying on internalized pre-training memory—is defined as Context Learning~\cite{dou2026clbench}.

Despite its critical importance, context learning remains a severe bottleneck for contemporary LLMs.
Recent evaluations on CL-Bench~\cite{dou2026clbench}, a comprehensive benchmark designed to assess this capability, reveal a striking mismatch between current model performance and real-world requirements. Frontier proprietary models, such as GPT-5.1, solve merely 23.7\% of complex context-dependent tasks, while state-of-the-art open-source models struggle even more, averaging around 13\% to 15\% in task-solving rates.

Addressing this context learning gap requires teaching models how to faithfully extract, internalize, and deduce from novel contexts. While recent advancements in synthetic data generation~\cite{wang2023selfinstruct, xu2025magpie,li2025glan,yu2025cotselfinstruct} and Chain-of-Thought (CoT) distillation~\cite{magister2023teaching, hsieh2023distilling, muennighoff2025s1, guha2025openthoughts, zhang2026quest} have successfully enhanced the reasoning capabilities of open-source models, directly applying these methods to context learning is highly problematic.
\begin{figure*}[h]
    \centering
    \includegraphics[width=\linewidth]{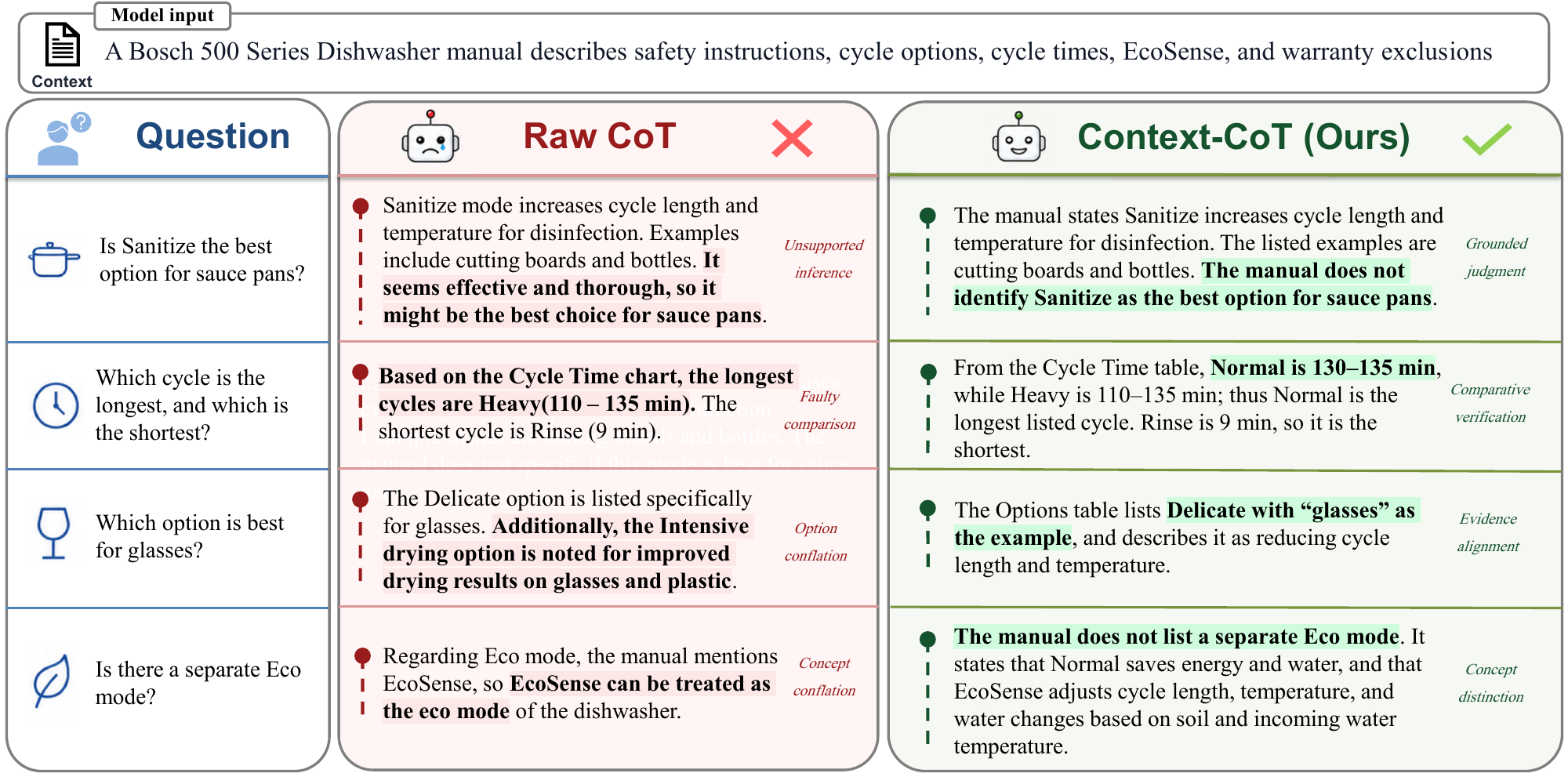}
    \caption{
    Qualitative comparison of reasoning trajectories generated by the baseline model (Raw CoT) and our fine-tuned model (Context-CoT) on a context-dependent question answering task. Given a novel context, such as a dishwasher manual , the baseline model frequently falls back on pre-trained priors, leading to unsupported assumptions, faulty comparisons, and concept confusion. In contrast, the model fine-tuned on our Context-CoT dataset explicitly extracts relevant textual evidence before reasoning, resulting in grounded judgments, accurate comparative verification, and strict evidence alignment.  
    }
    \label{fig:case_comparison}
\end{figure*}
The difficulty is threefold: teacher models may ignore the provided context and rely on parametric shortcuts, producing fluent but weakly grounded rationales~\cite{sun2026task, zhao2025understanding}; answer- or rubric-conditioned generation may lead to post-hoc rationalizations rather than context-derived reasoning~\cite{lewislim2025analysing, peng2026measuring}; and even correct teacher CoTs may be difficult for smaller student models to imitate effectively~\cite{li2025learningcommittee, he2026student, zhang2026quest}. These issues motivate a synthesis pipeline that jointly enforces contextual grounding, minimizes supervision leakage, and selects student-learnable reasoning trajectories.

To overcome these challenges,
we introduce \textbf{Context-CoT},
a novel data synthesis and fine-tuning framework designed explicitly to enhance context learning in open-source LLMs.
We propose a robust, three-stage CoT pipeline that generates high-quality, hallucination-free reasoning trajectories.
First, we employ \emph{Multi-Stage CoT Sampling}, which prompts strong generator models to explicitly extract and structure novel rules from the context before attempting to resolve the task, thereby grounding the reasoning process.
Second, we introduce \emph{Rubric-Based Minimum-Leakage Filtering} to reduce answer-conditioned rationalization in CoT synthesis. The key idea is to keep reference answers and full rubrics hidden from the teacher during generation, while using hidden rubric-based verification and minimal failed-rubric feedback to guide refinement only when necessary.
Finally, we apply \emph{Student-Aware CoT Selection} to identify reasoning trajectories that align best with the target open-source model's natural language distribution, ensuring that the synthesized knowledge is not only accurate but also highly assimilable.

To evaluate the effectiveness of Context-CoT, we conduct extensive experiments on CL-Bench~\cite{dou2026clbench}.
We fine-tune the Qwen3.5 model using approximately 4K high-quality training samples generated by our pipeline. Experimental results demonstrate that Context-CoT significantly outperforms strong baselines.
The improvements are consistent across various context-learning categories, with the most substantial gains observed in Domain Knowledge Reasoning.

Our contributions are summarized as follows:
\begin{itemize}
    \item We propose \textbf{Context-CoT},
    a novel three-stage framework to generate high-quality, hallucination-free reasoning trajectories for context learning.
    \item We construct a dedicated dataset of approximately 4K training samples,
    which explicitly enforces multi-step reasoning over novel contexts while minimizing reliance on static pre-trained knowledge.  
    \item Extensive evaluations on CL-Bench demonstrate the efficacy of our approach. Fine-tuning Qwen3.5 model on our synthesized dataset effectively improves the overall context-learning task-solving rate.
\end{itemize}

\section{Related Work}
\paragraph{Chain-of-Thought Data Synthesis.}
To enhance the reasoning capabilities of open-source models, recent methodologies have heavily relied on Chain-of-Thought (CoT) distillation \cite{hsieh2023distilling, feng2025cotevo, wu2025ded,jin2026dgpo} and synthetic data generation \cite{zelikman2022star, yu2025cotselfinstruct}.
While generating rationales from proprietary frontier models has proven effective for mathematical and logical reasoning \cite{shen2025mot}, synthesizing CoT for context learning presents unique challenges.
When presented with novel or niche long-tail content that largely extends beyond what models have acquired during pre-training, generator models frequently exhibit knowledge leakage \cite{baser2026thinkeval}.
Error analyses reveal that models often ignore the provided context or misuse it by defaulting to conflicting pre-trained priors.
Existing data filtering techniques \cite{yu2025cotselfinstruct} often struggle to distinguish between genuine context assimilation and hallucinated shortcuts.
To enable reliable evaluation and generation,
recent approaches have adopted fine-grained, task-level rubrics~\cite{baser2026thinkeval, rao2026autorubric,jin2026himac} designed as binary questions to verify solutions across multiple dimensions. Our work builds upon this paradigm by introducing a multi-stage synthesis pipeline that incorporates strict, rubric-based minimum leakage filtering and student-aware CoT selection, ensuring that the distilled reasoning trajectories are faithfully grounded in the provided context rather than pre-trained heuristics.
Extended related works are detailed in the appendix~\ref{sec:ext relat}.

\section{Methodology: The Context-CoT Pipeline}

The fundamental challenge of context learning is that models must acquire and reason over new knowledge—such as fictional rule systems, novel domain knowledge, or newly discovered empirical laws—that is often absent from or conflicts with their pre-trained parameters. Consequently, standard Chain-of-Thought (CoT) generation methods frequently fail, as generator models tend to hallucinate or fall back on their internal priors when faced with lengthy and unfamiliar contexts. To address this, we propose the Context-CoT Pipeline, a novel three-stage data synthesis framework designed to generate high-quality, hallucination-free, and highly learnable reasoning trajectories for open-source Large Language Models (LLMs).

\begin{figure*}[h]
    \centering
    \includegraphics[width=\linewidth]{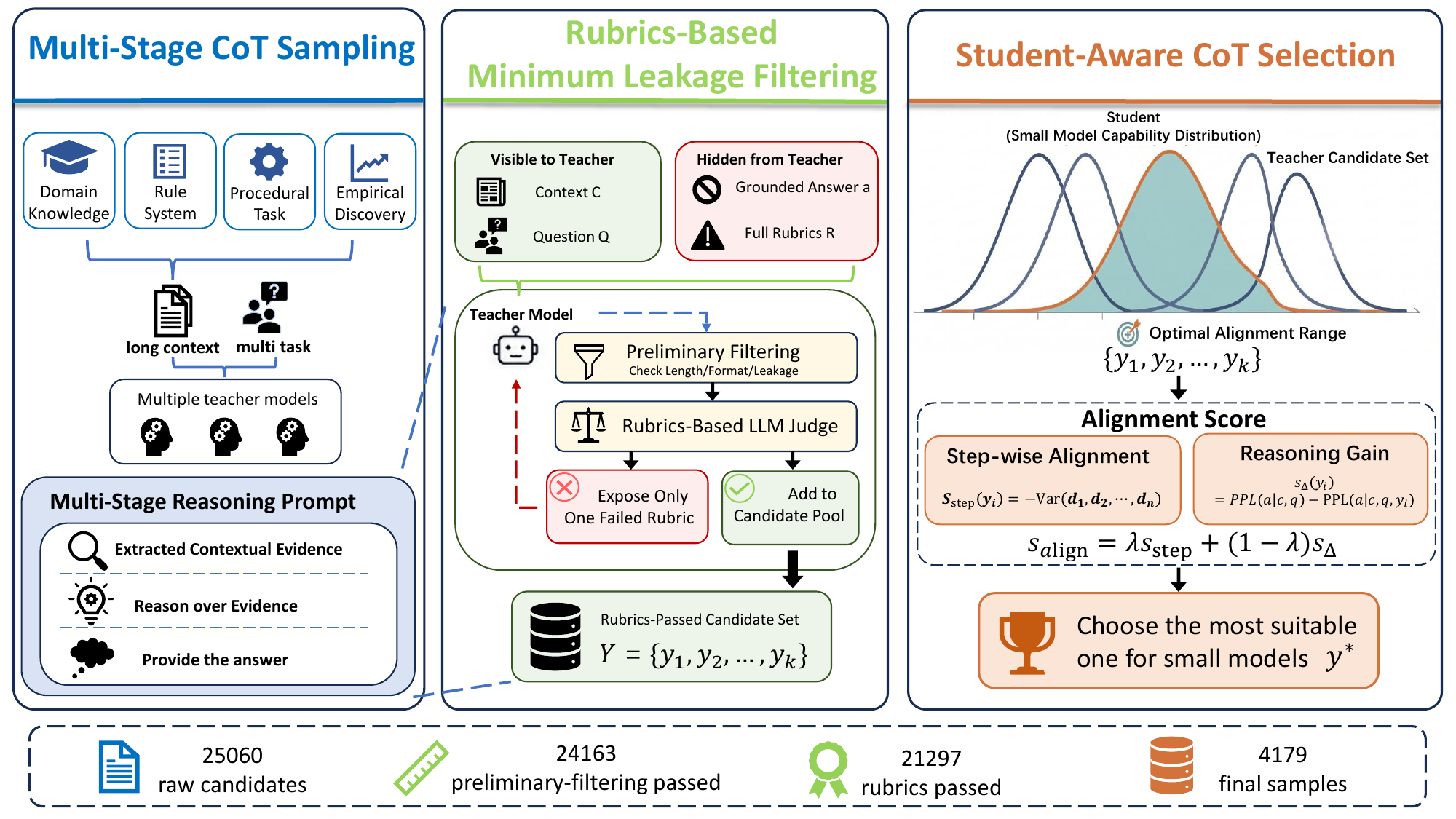}
    \caption{An overview of the Context-CoT data synthesis and filtering pipeline. The framework operates in three sequential stages: (1) Multi-Stage CoT Sampling, which prompts the teacher model to explicitly extract relevant contextual knowledge before reasoning; (2) Rubrics-Based Minimum Leakage Filtering, which utilizes preliminary length filtering and a rubric-based LLM judge to discard unfaithful reasoning trajectories ; and (3) Student-Aware CoT Selection, which evaluates the teacher candidate set to select the optimal reasoning trajectory ($y^{*}$) that best aligns with the target student model's capability distribution.}
    \label{fig:pipeline}
\end{figure*}

\subsection{Multi-Stage CoT Sampling}
Real-world context-dependent tasks are highly complex, requiring models to first identify relevant information within lengthy documents, which average 10.4K tokens in our benchmark setting, and subsequently apply it. To mirror this cognitive process and prevent the generator model from jumping to premature conclusions based on pre-trained biases, we decompose the CoT generation into a multi-stage sampling process. We prompt a highly capable frontier model to generate a two-phase reasoning trajectory. First, the model is explicitly instructed to scan the provided context and extract the specific rules, definitions, or procedural constraints relevant to the task before formulating an answer. This grounding step forces the model to anchor its reasoning in the provided text rather than its internal memory. Following this knowledge extraction, the model applies the isolated information to solve the task, employing either deductive reasoning for structured rule systems or inductive reasoning for discovering patterns in empirical data. We sample multiple reasoning paths for each task using a high temperature setting to ensure a diverse pool of candidate trajectories.

\subsection{Rubrics-Based Minimum-Leakage Filtering}

A key challenge in synthesizing CoT data for context learning is that a
seemingly correct rationale may be produced through leakage rather than
context-grounded reasoning. When the teacher model is given the reference answer or the full rubrics, it may generate a fluent post-hoc explanation that justifies the known answer instead of deriving it from the context. Such rationales are harmful because they teach the student model to imitate answer-conditioned explanations rather than to extract and apply novel contextual knowledge.

We therefore construct CoTs through an iterative minimum-leakage filtering process. For each context-question pair, 
we initialize the exposed rubric set as
\(\mathcal{R}^{(0)}_{\mathrm{exp}}=\emptyset\), so the teacher first receives
only the context and question. Each generated trajectory is filtered in two
steps. We first remove malformed outputs and trajectories with abnormal
lengths. The remaining candidates are then evaluated by a separate LLM judge
using the hidden reference answer and full rubric set, which verifies both
final-answer correctness and consistency between the answer and the reasoning
process.

Candidates that pass this verification are retained. For failed candidates, we
avoid exposing the full supervision signal. At refinement round \(t\), we
identify the failed rubric set \(\mathcal{F}^{(t)}\), expose only one failed
rubric \(r^{(t)}_{\mathrm{add}} \in \mathcal{F}^{(t)}\), and update
\[
\mathcal{R}^{(t+1)}_{\mathrm{exp}} =
\mathcal{R}^{(t)}_{\mathrm{exp}} \cup \{r^{(t)}_{\mathrm{add}}\}.
\]
The teacher then regenerates the CoT and answer under this minimally
strengthened prompt, and the new output is judged again with the full hidden
rubric set.
Before judging, we also apply a simple structural filter to remove
malformed trajectories with abnormal lengths.

This procedure exposes only the minimum additional information needed to repair
failed reasoning, while keeping the reference answer hidden throughout
generation. Compared with answer-conditioned CoT synthesis, iterative
minimum-leakage filtering encourages the teacher to first solve the task from
the context itself and only uses rubrics as targeted corrective signals when
necessary. As a result, the retained CoTs are more likely to encode the desired
context-learning behavior: extracting relevant contextual information,
reasoning over it, and producing an answer that satisfies fine-grained task
constraints.

\subsection{Student-Aware CoT Selection}

Although rubric-based filtering ensures the correctness and faithfulness of a
candidate CoT, it does not guarantee that the trajectory is suitable for the
target student model to imitate. In practice, different teacher models may generate correct
CoTs with different reasoning granularities and linguistic styles, which can
introduce additional alignment costs for a smaller student model.

To select the most learnable trajectory, we introduce a student-aware alignment
score. For each context-question pair $(c,q)$, let
$\mathcal{Y}_{r}(c,q)=\{y_1,y_2,\dots,y_k\}$ denote the set of CoT candidates
that pass rubric-based filtering.
For notation simplicity, we write
$\mathcal{Y}=\mathcal{Y}_{r}(c,q)$ when $(c,q)$ is clear from context.
For each candidate $y_i$, we compute two student-aware scores: a step-wise alignment score $S_{\mathrm{step}}$, which measures whether the reasoning difficulty is smoothly distributed across reasoning steps, and a reasoning-gain score $S_{\Delta}$, which measures whether conditioning on the CoT reduces the student's uncertainty about the final answer.

Since the two scores may have different numerical scales, we normalize them
within the candidate set $\mathcal{Y}_{r}(c,q)$ as
\begin{equation}
\widetilde{S}_{m}(y_i)
=
\frac{
S_{m}(y_i)-S_m^{\min}
}{
S_m^{\max}-S_m^{\min}+\epsilon
},
\quad
m \in \{\mathrm{step}, \Delta\},
\label{eq:score_normalization}
\end{equation}
where $\epsilon$ is a small constant for numerical stability.

The final student-aware alignment score is then defined as
\begin{equation}
S_{\mathrm{align}}(y_i)
=
\lambda \widetilde{S}_{\mathrm{step}}(y_i)
+
(1-\lambda)\widetilde{S}_{\Delta}(y_i).
\label{eq:student_aware_alignment}
\end{equation}
All component scores are computed with respect to the target student model
$\mathcal{M}_{s}$, which is omitted from the notation for brevity. Here,
$\lambda \in [0,1]$ controls the trade-off between step-wise alignment and
reasoning gain.

Specifically, we split each CoT into reasoning steps
$y_i=\{s_1,s_2,\dots,s_n\}$ and compute the student-model difficulty of each
step as
\begin{equation}
d_j
=
-\frac{1}{|s_j|}
\sum_{t=1}^{|s_j|}
\log p_{\mathcal{M}_{s}}
\left(
s_{j,t}
\mid
c, q, s_{<j}, s_{j,<t}
\right).
\label{eq:step_difficulty}
\end{equation}

The step-wise alignment score is defined as
\begin{equation}
S_{\mathrm{step}}(y_i)
=
-\operatorname{Var}(d_1,d_2,\dots,d_n),
\label{eq:step_score}
\end{equation}
which penalizes difficulty jumps between steps.

The reasoning-gain score is defined as
\begin{equation}
S_{\Delta}(y_i)
=
\mathrm{PPL}_{\mathcal{M}_{s}}(a \mid c, q)
-
\mathrm{PPL}_{\mathcal{M}_{s}}(a \mid c, q, y_i),
\label{eq:delta_score}
\end{equation}

where $a$ is the reference answer. A positive $S_{\Delta}$ indicates that the CoT helps reduce the student's uncertainty about the answer.

Finally, among all candidates, we select the CoT with the highest
student-aware alignment score:
\begin{equation}
y^{*}
=
\arg\max_{y_i \in \mathcal{Y}_{r}(c,q)}
S_{\mathrm{align}}(y_i; \mathcal{M}_{s}).
\label{eq:cot_selection}
\end{equation}

\section{Dataset Construction \& Analysis}

\subsection{Overview and Design Goals}

A central obstacle to improving context learning is the lack of training data that simultaneously requires long-context dependency, multi-step reasoning, and low reliance on pre-trained knowledge. Existing long-context question answering datasets often emphasize retrieval or reading comprehension, while reasoning-oriented datasets usually involve shorter contexts and may be partially solved from parametric knowledge. Thus, directly fine-tuning on these datasets does not necessarily teach models to acquire and apply novel knowledge from a provided context.

\begin{table*}[h]
    \centering
    \scriptsize
    \setlength{\tabcolsep}{4pt}
    \resizebox{\textwidth}{!}{
    \begin{tabular}{lccccc}
        \toprule
        Dataset / Paradigm
        & Context-grounded
        & CoT
        & Fine-grained Sup.
        & Leakage Control
        & Student-specific Subset \\
        \midrule
        Self-Instruct~\citep{wang2023selfinstruct}
        & $\times$ & $\times$ & $\times$ & $\times$ & $\times$ \\
        MetaMathQA~\citep{yu2023metamath}
        & $\times$ & $\checkmark$ & $\times$ & $\times$ & $\times$ \\
        PRM800K~\citep{lightman2023let}
        & $\times$ & $\checkmark$ & $\checkmark$ & $\times$ & $\times$ \\
        LongAlign~\citep{bai2024longalign}
        & $\checkmark$ & $\times$ & $\times$ & $\times$ & $\times$ \\
        OpenThoughts~\citep{guha2025openthoughts}
        & $\times$ & $\checkmark$ & $\checkmark$ & $\times$ & $\times$ \\
        LongFaith~\citep{yang2025longfaith}
        & $\checkmark$ & $\checkmark$ & $\checkmark$ & $\times$ & $\times$ \\

        \midrule
        \textbf{Context-CoT}
        & $\checkmark$ & $\checkmark$ & $\checkmark$ & $\checkmark$ & $\checkmark$ \\
        \bottomrule
    \end{tabular}
    }
    \caption{Comparison with representative training data construction paradigms.}
    \label{tab:dataset_comparison}
\end{table*}

To address this gap, we construct \textbf{Context-CoT}, a synthetic training dataset designed for context learning. Each instance consists of a long context, a context-dependent question, a reference answer, fine-grained evaluation rubrics, and one selected CoT trajectory. We generate fictional, modified, or long-tail contexts and design questions that require models to extract and apply context-specific rules, definitions, procedures, or empirical patterns.

To avoid evaluation contamination, we do not use any contexts, questions, answers, or rubrics from the CL-Bench. All training instances are independently generated and only follow the broad task taxonomy and evaluation motivation of CL-Bench.

\subsection{Dataset Construction}
We construct the training data through a staged context-task generation and CoT
selection pipeline. We first generate long context-task instances covering four
context-learning categories: Domain Knowledge Reasoning, Rule System
Application, Procedural Task Execution, and Empirical Discovery \& Simulation.
To improve diversity, the generation prompts vary topic keywords, task
categories, domain styles, and few-shot exemplars.
\begin{table}[h]
    \centering
    \small
    \setlength{\tabcolsep}{4pt}
    \begin{tabular}{lcc}
        \toprule
        Stage & Trajectories & Retention Rate\\
        \midrule
Teacher candidates & 25,060 & 100.00\%\\
Length-filtered candidates & 24,163 & 96.42\%\\
Rubric-passed candidates & 21,297 & 84.98\%\\
Final selected samples & 4,179 & 16.68\%\\
        \bottomrule
    \end{tabular}
    \caption{Statistics of the Context-CoT filtering pipeline.}
    \label{tab:filtering_statistics}
\end{table}

For each generated context, we further generate multiple questions, reference answers, and fine-grained evaluation rubrics. Each question is designed to require information from the corresponding context, while each rubric serves as a binary verification criterion for checking whether a model output satisfies a specific factual, procedural, computational, or formatting requirement.

Based on these context-task instances, we sample CoT trajectories from strong teacher models, using their reasoning configurations. The raw teacher-generated trajectories are not directly used for fine-tuning. Instead, they are passed through the proposed filtering and selection pipeline: preliminary filtering removes abnormal trajectories, rubric-based filtering retains candidates that satisfy fine-grained context-specific requirements, and student-aware selection chooses the trajectory that is best aligned with the target student model.

\subsection{Data Analysis}

Table~\ref{tab:filtering_statistics} summarizes the statistics of the constructed
training data. The dataset contains 705 long contexts, 4,179 question-answer
pairs, and 30,553 fine-grained rubrics, covering four types of context-learning
tasks. On average, each context is paired with 5.93 questions, and each question
is associated with 7.31 rubrics. Starting from 25,060 raw teacher-generated CoT
candidates, the filtering and selection pipeline finally yields 4,179 selected
training samples, with one CoT trajectory selected for each question-answer
pair.

We compare Context-CoT with representative training datasets in
Table~\ref{tab:dataset_comparison}, covering instruction tuning, reasoning-data construction and long-context alignment. Existing datasets typically focus on one or a few aspects of training data construction. For example, LongFaith also synthesizes faithful long-context reasoning data, but its reasoning chains are generated with ground-truth and citation-based guidance, whereas Context-CoT keeps the reference answer hidden during CoT generation and exposes only minimal corrective rubric information when refinement is needed. In addition, Context-CoT further performs student-aware selection to construct a training subset tailored to the target student model.

\section{Experiments}

\subsection{Experimental Setup}

\begin{table*}[t]
    \centering
    \small
    \setlength{\tabcolsep}{4.5pt}
    \begin{tabular}{llcccccc}
        \toprule
        model & Method & Overall & $\Delta$ & Domain & Empir. \& Sim. & Proc. & Rule \\
        \midrule
        Qwen3.5-4B
        & Base Model 
        & 9.06 & -- & 9.63 & 11.05 & 6.91 & 9.40 \\
        & Answer-only SFT 
        & 9.32 & +0.26 & 10.87 & 8.38 & 7.66 & 9.14 \\
        & Answer-exposed CoT SFT 
        & 8.59 & -0.47 & 9.66 & 9.52 & 7.23 & 8.08 \\
        & \textbf{Context-CoT SFT}
        & \textbf{12.85} & \textbf{+3.79} 
        & \textbf{14.88} & \textbf{12.31} & \textbf{10.25} & \textbf{12.72} \\
        \midrule
        Llama3.2-3B
        & Base Model 
        & 3.04 & -- & 3.44 & 2.59 & 3.29 & 2.52 \\
        & Answer-only SFT 
        & 4.25 & +1.21 & 4.19 & 5.18 & 2.91 & 5.02 \\
        & Answer-exposed CoT SFT 
        & 2.71 & -0.33 & 2.22 & 2.21 & 3.19 & 3.08 \\
        & \textbf{Context-CoT SFT}
        & \textbf{7.57} & \textbf{+4.53} 
        & \textbf{9.11} & \textbf{5.76} & \textbf{7.89} & \textbf{6.20} \\
        \bottomrule
    \end{tabular}
    \caption{Main experimental results on CL-Bench.}
    \label{tab:main_results}
\end{table*}

\paragraph{Tasks and Datasets.}
We use the CL-Bench test set as our main benchmark, which contains 1,899 long-context tasks across four context-learning categories: Domain Knowledge Reasoning (Domain), Empirical Discovery \& Simulation (Empir. \& Sim.), Procedural Task Execution (Proc.), and Rule System Application (Rule). Each task is evaluated with multiple fine-grained rubrics, and a task is counted as correct only if all rubrics are satisfied by the judge LLM. We evaluate two open-source student models: Qwen3.5-4B as the primary model for detailed analysis, and Llama3.2-3B as a smaller model for cross-model validation. For fairness, all SFT-based methods are built from the same underlying context-question-answer pool and use the same training scale.

\paragraph{Baselines.}
We compare Context-CoT with the following baselines:
(i) \textit{Base Model}: the original student model without task-specific fine-tuning.
(ii) \textit{Answer-only SFT}: supervised fine-tuning directly on the generated context-question-answer triples. This baseline uses the reference answers from our constructed training instances as the only supervision signal, without using any teacher-generated CoT trajectories or any subsequent CoT filtering and selection stages.
(iii) \textit{Answer-exposed CoT SFT}: a standard teacher-distillation baseline where the teacher is given the context, question, and reference answer during CoT generation, and the generated CoT trajectories are then used for SFT. This baseline is designed to test whether directly distilling answer-conditioned rationales is sufficient for context learning. In our implementation of Answer-exposed CoT SFT, we use DeepSeek-R1 as the teacher model.

All fine-tuning experiments use the same training budget, including the same number of training epochs, learning rate, maximum input length, and effective batch size, to ensure a fair comparison among different data construction methods.

\subsection{Experimental Results}

Table~\ref{tab:main_results} shows that Context-CoT consistently improves the context-learning performance of open-source student models. On the primary Qwen3.5-4B model, Context-CoT improves the overall CL-Bench score from 9.06\% to 12.85\%, yielding a 3.79-point absolute gain. In contrast, Answer SFT brings only a slight improvement, and Answer-exposed CoT SFT even underperforms the base model. This suggests that simply exposing the reference answer to a strong teacher during CoT generation may produce fluent but potentially harmful rationales, rather than teaching the student how to derive answers from the provided context.

On Llama3.2-3B, Context-CoT also improves over the corresponding base model, indicating that the proposed pipeline is not tied to a single student model. Although smaller open-source models still lag behind stronger frontier thinking models reported by CL-Bench, these results suggest that high-quality context-grounded CoT supervision can improve their ability to extract, internalize, and apply novel contextual information.

\paragraph{Statistical testing.}
For the primary Qwen3.5-4B, we conduct paired bootstrap resampling and McNemar's exact test to assess the stability of the observed improvement. Context-CoT improves over the base model from 9.06\% to 12.85\%, yielding a 3.79-point absolute gain with a 95\% confidence interval of [1.68, 5.03]. The gain is also statistically significant under McNemar's exact test ($b=83$, $c=143$, $p=7.95\times10^{-5}$), where $b$ denotes tasks solved only by the base model and $c$ denotes tasks solved only by Context-CoT. These results suggest that the improvement is consistent across evaluation instances rather than being driven by a small number of unstable cases.

More specifically, Answer-only SFT brings only marginal gains, indicating that directly training on answers provides limited benefits for context learning. However, Answer-exposed CoT SFT shows an overall performance degradation. This result supports our motivation that answer-conditioned CoT generation may introduce post-hoc rationalization or distribution mismatch: the rationale can explain a known answer, but may not reflect a context-derived reasoning process that the student can reliably learn. In contrast, Context-CoT achieves a much larger improvement, with the largest gain in Domain Knowledge Reasoning and clear improvements in both Procedural Task Execution and Rule System Application. We further analyze these improvements in the following sections.

\subsection{Ablation Study}

\paragraph{Ablation of Pipeline Modules.}
We ablate the three major components of Context-CoT and fine-tune the student model under the same training settings. As shown in Table~\ref{tab:pipeline_ablation}, removing any component leads to performance degradation, showing that Context-CoT benefits from the joint design of candidate generation, leakage-aware filtering, and student-aware selection.
\begin{table*}[h]
    \centering
    \begin{tabular}{lccccc}
        \toprule
        Method & Overall & Domain & Empir. \& Sim. & Proc. & Rule \\
        \midrule
        Full Context-CoT & 12.85 & 14.88 & 12.31 & 10.25 & 12.72 \\
        w/o Multi-stage Answering & 12.14 & 14.12 & 11.28 & 9.57 & 12.01 \\
        w/o Minimum-leakage Filtering & 11.07 & 13.73 & 10.47 & 8.41 & 10.22 \\
        w/o Student-aware Alignment & 11.38 & 14.06 & 10.88 & 9.03 & 10.27 \\
        \bottomrule
        \\[-0.8em]
    \end{tabular}
     \caption{Ablation study of different modules in the Context-CoT pipeline.}
      \label{tab:pipeline_ablation}
\end{table*}
\begin{figure}[h]
  \centering
  \includegraphics[width=\linewidth]{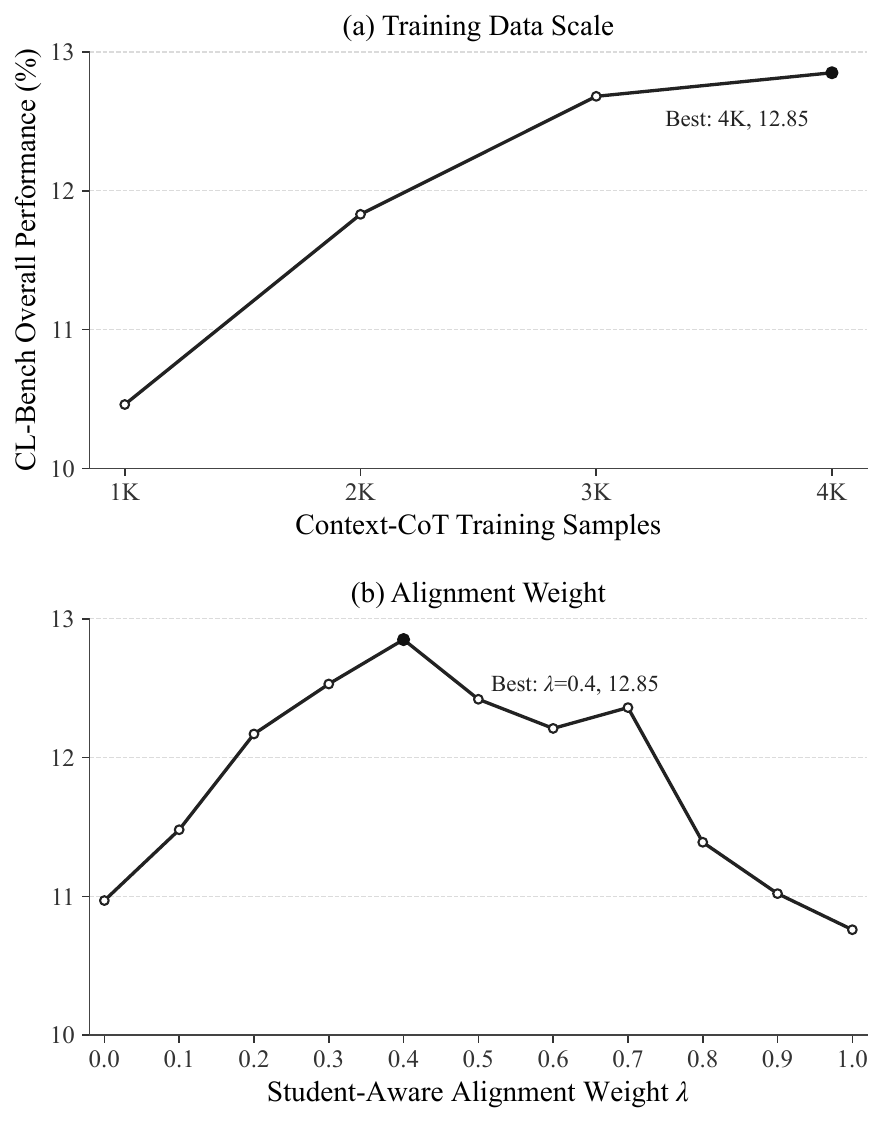} 
  \caption{Ablation studies on training data scale and the student-aware alignment
weight $\lambda$. (a) Increasing the number of Context-CoT training samples
consistently improves performance, with diminishing gains after 3K samples.
(b) The model achieves the best performance when $\lambda=0.4$, indicating that
balancing step-wise alignment and reasoning gain is important.}
  \label{fig:lambda_ablation}
\end{figure}

Removing minimum-leakage filtering causes the largest drop, indicating that plausible CoTs may contain answer-conditioned rationalization or pre-training shortcuts if they are not verified against hidden context-specific rubrics. Removing student-aware alignment selection also substantially hurts performance, suggesting that correct and context-faithful CoTs are not necessarily the most learnable ones for the target student model. Removing multi-stage answering leads to a smaller drop, showing that explicit context extraction improves the initial candidate pool, while later filtering can partially compensate for weaker generation.

Overall, these results suggest that the effectiveness of Context-CoT comes from selective synthesis rather than CoT generation alone: the pipeline must generate diverse rationales, filter leakage-prone ones, and select trajectories that the student can reliably learn from.

\paragraph{Training Data Scale Ablation}
To study how the amount of synthesized training data affects context-learning performance, we further fine-tune the student model with different numbers of Context-CoT training samples. As shown in Figure~\ref{fig:lambda_ablation}(a), performance improves monotonically as the training set size increases from 1K to 4K samples. The gain is substantial when increasing the data size from 1K to 3K, while the improvement becomes smaller from 3K to 4K, suggesting a diminishing-return trend under the current model and training setting.

These results indicate that Context-CoT benefits from larger amounts of
filtered CoT supervision, but the improvement is not merely due to adding more training examples. Even with 1K samples, the model already outperforms the base model, while the full 4K dataset achieves the best performance. This suggests that the proposed filtering and selection pipeline can produce effective training signals even at relatively small data scales.

\paragraph{Hyperparameter Ablation}
In Section 3.3, our proposed alignment score contains a hyperparameter $\lambda$, which controls the weight between a step-wise alignment score $S_{\mathrm{step}}$ and a reasoning-gain score $S_{\Delta}$. To further explore the influence of this hyperparameter, we conduct a comprehensive analysis and visualize the results in Figure~\ref{fig:lambda_ablation}(b).

We find that $\lambda=0.4$ yields the best performance, suggesting that Context-CoT benefits from balancing step-wise alignment and reasoning gain. Performance remains strong when $\lambda \in [0.2,0.7]$, but drops at the extremes. This indicates that relying only on reasoning gain or step-wise smoothness is suboptimal for selecting learnable CoTs.

\section{Conclusion}

In this work, we present Context-CoT, a novel data synthesis and fine-tuning framework that enhances context learning in open-source LLMs by transitioning them from basic prompt followers to robust context learners. We construct a high-quality, context-grounded dataset of approximately 4K samples via a three-stage pipeline comprising multi-stage evidence sampling, rubric-based minimum-leakage filtering, and student-aware perplexity alignment.
Extensive evaluations on CL-Bench demonstrate that open-source models fine-tuned with Context-CoT achieve substantial and statistically significant performance gains, particularly in complex Domain Knowledge Reasoning. These results validate that data quality, strict contextual grounding, and student-centric alignment are paramount for expanding the dynamic knowledge-scaling capabilities of LLMs.

\section*{Limitations}
Although Context-CoT demonstrates significant efficacy in mitigating context-neglect errors, our current framework and empirical evaluations are strictly constrained to text-based tasks. We deliberately scoped our investigation to text as it serves as the foundational testbed for complex reasoning synthesis; however, real-world information assimilation is inherently multimodal. A highly promising trajectory for future research is extending this context-grounded distillation pipeline to Multimodal Large Language Models (MLLMs). Given that context neglect is notably exacerbated in dynamic and continuous visual streams, adapting our multi-stage evidence extraction and minimum-leakage filtering mechanisms to tackle temporal reasoning and memory modeling in ultra-long video content represents a formidable yet critical next step in advancing robust context learning.



\bibliography{custom}

\appendix

\clearpage
\section{Extended Related Work}
\label{sec:ext relat}

\subsection{In Context Learning}
Prompt engineering \cite{schulhoff2025promptreport} and In-Context Learning (ICL) \cite{dong2024surveyicl} have traditionally enabled language models to infer task formats and expected behaviors through carefully designed instructions and few-shot demonstrations \cite{brown2020gpt3}.
However, these paradigms primarily elicit reasoning over prompts using static, pre-trained knowledge.
In contrast, real-world applications increasingly rely on context engineering, such as Retrieval-Augmented Generation (RAG) \cite{lewis2020rag, sharma2025ragsurvey, zhao2024ragaigc} and agentic pipelines \cite{yao2023react, schick2023toolformer}, to provide models with novel, task-specific information.
Despite advancements in organizing and retrieving context from diverse sources like private documents and databases,
existing research has largely overlooked the foundational capability of context learning \cite{dou2026clbench}—whether models can genuinely acquire and apply new knowledge from complex, provided contexts to reason beyond their pre-trained parameters.

\subsection{Long-Context Reasoning and Instruction Following}

Context learning is closely related to long-context reasoning and instruction following, as complex real-world tasks often involve lengthy documents with intricate constraints \cite{liu2023lost, kuratov2024babilong}.
A variety of benchmarks \cite{bai2025longbenchv2, hsieh2024ruler} have been proposed to evaluate models in long-context settings, focusing on tasks like document question answering \cite{wang2024loong}, summarization \cite{laban2024summhay}, and retrieval-in-a-haystack \cite{kuratov2024babilong}. Similarly, verifiable instruction-following benchmarks assess models' adherence to specific formatting or compositional constraints \cite{pyatkin2025ifbench, zou2025eifbench}.
Nevertheless, these evaluations primarily test reading comprehension, retrieval, or shallow heuristics on relatively simple tasks \cite{modarressi2025nolima}.
Context learning demands far richer knowledge acquisition, requiring models to internalize vertical domain knowledge, deduce from novel rule systems, or induce laws from empirical data.
Consequently, strong performance on long-context or instruction-following tasks constitutes a necessary but entirely insufficient condition for effective context learning.

\section{Discussion and Analysis}

\subsection{Why Context-CoT Improves Context Learning}

The experimental results suggest that the improvement of Context-CoT mainly comes from two aspects. First, the minimum-leakage construction process encourages the teacher model to derive the answer from the provided context rather than explaining a known reference answer. This reduces the risk of training the student model on rationales that rely on answer leakage or pre-trained shortcuts. Second, the student-aware alignment stage selects reasoning trajectories that are more compatible with the target model's own distribution. As a result, the student model can more easily imitate the reasoning process and internalize useful context-dependent reasoning patterns during fine-tuning.

\subsection{Category-wise Performance Analysis}

After fine-tuning, we observe that the model mainly improves on Domain Knowledge Reasoning and Procedural Task Execution, while the gains on Empirical Discovery \& Simulation and Rule System Application are relatively smaller. We believe this difference is related to the intrinsic nature of these task categories.

For domain knowledge reasoning and procedural task execution, the teacher-generated CoT demonstrations provide learnable reasoning patterns, such as how to identify relevant contextual facts, decompose the task, and follow a procedure step by step. These patterns can be effectively transferred to the student model through supervised fine-tuning, leading to clear improvements in answer quality.

In contrast, empirical discovery and rule system application require deeper context learning abilities. Empirical discovery tasks often require models to abstract hidden laws from a large number of observations or simulated physical-world phenomena. Rule system application tasks may involve many interacting rules, including priority conflicts and state-dependent decisions. These tasks are therefore less likely to be solved by learning surface-level reasoning forms alone. This suggests that such categories may require stronger training signals beyond static CoT imitation, such as explicit rule-state tracking, interactive reasoning, or online self-improvement.

\subsection{Limitations and Future Work}

Although the Context-CoT pipeline improves performance on CL-Bench, it still has several limitations.

\paragraph{High Construction Cost.} The construction process is relatively expensive. Our filtering process relies on sampling a large number of diverse CoT candidates from multiple strong teacher models. Compared with standard CoT construction, this introduces additional computational and API costs.

\paragraph{Offline Distillation Framework.} Our current framework is fully offline. The CoT data are selected from a pre-generated candidate pool, and the student model does not interact with the environment or dynamically influence the data construction process. While this stabilizes training, it may limit the alignment between the evolving student model and the selected reasoning trajectories. In future work, we plan to extend Context-CoT to an online distillation framework.

\paragraph{Dependence on Rubric-based Judgments.} Our filtering process depends on rubric-based LLM judgment. Although fine-grained binary rubrics reduce evaluation ambiguity, judge errors may still affect candidate selection. Future work can explore more robust verification methods, such as tool-assisted checking or multi-judge agreement.

\section{Training Details}

All fine-tuning experiments are conducted on Qwen3.5-4B-Base using supervised
fine-tuning (SFT) with LoRA adapters. LoRA is applied to the attention
projection layers and MLP projection layers, including
\texttt{q\_proj}, \texttt{k\_proj}, \texttt{v\_proj}, \texttt{o\_proj},
\texttt{gate\_proj}, \texttt{up\_proj}, and \texttt{down\_proj}. We use rank
\(r=32\), LoRA alpha \(=64\), and dropout \(=0.05\).

We use LoRA to provide a parameter-efficient and compute-controlled SFT setting.
All SFT-based baselines and our method use the same LoRA configuration, so the
comparison isolates the effect of data construction rather than differences in
fine-tuning capacity or optimization budget.

For the main Context-CoT training runs, we train for 1 epoch with a per-device
batch size of 1 and gradient accumulation steps of 16, resulting in an effective
batch size of 16 on a single GPU. The learning rate is \(1\times10^{-5}\), the
optimizer is AdamW implemented as \texttt{adamw\_torch}, the warmup ratio is
0.05, and the weight decay is 0.01. We use a cosine learning-rate scheduler,
bfloat16 training, and gradient checkpointing.

We split the constructed training set into 95\% training and 5\% validation
examples. During optimization, we apply a loss mask to all prompt tokens, including the
context, question, and system/user instructions. The language modeling loss is
computed only on the assistant response tokens used as the SFT target.

For fairness, all fine-tuned baselines use the same training budget, including
the same base model, number of epochs, learning rate, batch size, effective
batch size, maximum sequence length, precision, and optimization settings. This
ensures that comparisons reflect differences in training data construction
rather than differences in optimization budget.

The evaluation protocol follows the main text. After fine-tuning, we merge the
LoRA adapters into the base model and generate outputs on the CL-Bench test set
using vLLM. The merged fine-tuned model is decoded with temperature \(0.6\),
top-\(p=0.95\), and top-\(k=20\). Predictions are evaluated by an LLM judge
under the rubric-based CL-Bench protocol. The judge model is served with vLLM and
decoded greedily with temperature \(0.0\).

\section{Multi-Teacher Sampling Details}

To construct the candidate reasoning pool, we perform multi-teacher sampling
using an OpenAI-compatible API backend. We use the asynchronous client interface
and query six teacher models: minimax-m2.5, qwen3.5-plus, doubao-seed-2.0-pro, deepseek-v3.2, kimi-k2.5 and glm-4.7. For each context-question instance, we sample one candidate response from each
teacher model, yielding one candidate per teacher.

All teacher models are queried with the same base decoding configuration unless
the API requires model-specific reasoning controls. We use temperature \(0.7\),
top-\(p=0.95\), and a maximum generation budget of 24{,}384 tokens to allow
complete long-form reasoning responses. Responses are generated in streaming
mode and processed asynchronously with a concurrency limit of 8 requests.

\section{Filtering and Selection Implementation Details}

This section provides additional implementation details for the filtering and
student-aware selection stages. The filtering stage contains two parts: a
lightweight structural filter that removes malformed generations, followed by a
hidden rubric-based judge that verifies answer correctness and reasoning
faithfulness.

\subsection{Preliminary Structural Filtering}

Before applying the hidden rubric judge, we first remove candidates that are not
suitable as supervised fine-tuning targets. Specifically, we discard outputs that
cannot be parsed into a separable reasoning trajectory and final answer, contain
a missing or degenerate reasoning field, contain no valid final answer, are
truncated, or explicitly refer to hidden supervision signals such as reference
answers, rubrics, or oracle judgments.

This step is applied before semantic judging and is used only to remove
structurally invalid or obviously leakage-prone generations. In our construction,
the preliminary filter reduces the raw candidate pool from 25,060 trajectories to
24,163 trajectories, removing 897 malformed or unusable candidates. The remaining
candidates are then passed to the hidden rubric-based judge.

\subsection{Minimum-Leakage Refinement}

After preliminary filtering, each candidate is evaluated by a hidden
rubric-based judge. The judge receives the reference answer and the full rubric
set, while the teacher model does not observe either of them during initial CoT
generation. A candidate is added to the rubric-passed candidate pool only if the
judge determines that both the final answer and the reasoning process satisfy
all context-specific rubrics.

If a candidate fails verification, we perform limited minimum-leakage refinement.
Instead of exposing the reference answer or the full rubric set, we expose only
one failed rubric as a targeted correction signal and ask the teacher to
regenerate the reasoning trajectory and final answer. We allow at most five
refinement rounds for each candidate. Therefore, a teacher candidate can receive
at most five failed-rubric hints during generation, while the reference answer
remains hidden throughout the entire process. Candidates that still fail after
the final refinement round are discarded.

This refinement protocol provides localized feedback for repairing specific
reasoning errors without turning CoT generation into answer-conditioned
rationalization. In the final candidate pool, only trajectories that pass all
hidden rubrics are retained for student-aware selection.

\subsection{Student-Aware Step Segmentation}

For student-aware selection, each rubric-passed CoT is segmented into reasoning
steps before computing the step-wise alignment score. We first split the
trajectory using explicit discourse markers such as numbered steps, bullet
points, or step-level transitions. If such markers are absent, we use paragraph
boundaries as a fallback. This produces a sequence of reasoning steps
\(\{s_1,\dots,s_n\}\).

For each step, we compute its average negative log-likelihood under the target
student model conditioned on the context, question, and previous reasoning
steps. The resulting step-difficulty sequence is used to compute the
variance-based step-wise alignment score. Intuitively, this score favors
trajectories whose reasoning difficulty is distributed smoothly across steps,
rather than trajectories with abrupt inferential jumps that may be difficult for
a smaller student model to imitate.

\subsection{Candidate-Level Normalization and Selection}

For each context-question pair, student-aware selection is applied only among
candidates that have passed rubric-based filtering. We compute both the
step-wise alignment score and the reasoning-gain score for each candidate, and
normalize the two scores within the candidate set before combining them. This
within-instance normalization avoids comparing raw score magnitudes across
different questions or contexts.

After normalization, we compute the final student-aware alignment score using
the same \(\lambda=0.4\) setting as in the main experiment. If only one
rubric-passed candidate is available for a context-question pair, we select it
directly. Otherwise, we select the candidate with the highest combined alignment
score. The final SFT dataset contains one selected reasoning trajectory for each
context-question instance, yielding 4,179 training samples.

\section{Analysis of Student-Aware Selection Principles}

We further analyze two single-principle variants of the student-aware CoT selection stage on Qwen3.5-4B. These variants correspond to two different selection principles: selecting CoTs that most help the student predict the final answer, and selecting CoTs whose reasoning difficulty is most smoothly distributed across steps.

Reasoning-gain selection chooses the candidate CoT that most improves the student model's prediction of the final answer. Concretely, it selects candidates according to $\widetilde{S}_{\Delta}$ only:

\[
y^{*}_{\Delta}
=
\arg\max_{y_i \in \mathcal{Y}_{r}(c,q)}
\widetilde{S}_{\Delta}(y_i;\mathcal{M}_{s}).
\]

Since $\mathrm{PPL}_{\mathcal{M}_{s}}(a \mid c,q)$ is fixed for all candidates of the same instance, this is equivalent to selecting the rubric-passed CoT that minimizes $\mathrm{PPL}_{\mathcal{M}_{s}}(a \mid c,q,y_i)$. This strategy
therefore favors answer-level utility: the selected trajectory is the one that most helps the student model predict the reference answer. This variant achieves an overall CL-Bench score of 10.97, outperforming the base model but remaining below the full student-aware selection objective.

Step-alignment selection follows a different principle. It selects candidates according to $\widetilde{S}_{\mathrm{step}}$ only:

\[
y^{*}_{\mathrm{step}}
=
\arg\max_{y_i \in \mathcal{Y}_{r}(c,q)}
\widetilde{S}_{\mathrm{step}}(y_i;\mathcal{M}_{s}).
\]

This strategy favors CoTs whose reasoning difficulty is smoothly distributed across steps. It focuses on the learnability of the reasoning path itself, rather than how strongly the path reduces the student's uncertainty about the final answer. This variant achieves an overall score of 10.65, also improving over the base model but underperforming the full objective.

These results suggest that both answer-level utility and step-level learnability provide useful training signals, but neither criterion alone is sufficient. Effective CoT selection should favor trajectories that both help the student infer the answer and present a reasoning path whose difficulty is compatible with the student model.

\section{Prompt Templates}

This appendix summarizes the main prompt templates used for constructing
Context-CoT. We report representative templates rather than every instantiated
prompt, since concrete prompts are obtained by substituting category-specific
variables such as subcategory, topic seeds, style constraints, and length
requirements. The prompt design covers four stages: context construction,
system-instruction construction, task and rubric construction, and
minimum-leakage CoT synthesis. We also report the answer-exposed CoT baseline
prompt for comparison.

\subsection{Context Construction Prompts}

We generate long contexts for four types of context-learning tasks: Domain
Knowledge Reasoning, Rule System Application, Procedural Task Execution, and
Empirical Discovery \& Simulation. Each prompt instantiates a category-specific
template with sampled variables such as subcategory, concrete focus, topic seeds,
and minimum length. The generated contexts are required to be self-contained,
internally consistent, and dense enough to support multi-step reasoning.

\paragraph{Domain knowledge context generation.}
The following template is used to generate contexts that introduce novel domain
knowledge inside the document itself.

\begin{Prompt}
# Role
You are a benchmark context designer for Context-CoT. Your job is to create a
realistic, fully invented professional document that teaches novel domain
knowledge inside the context itself.

# Target Category
CL-Bench category: Domain Knowledge Reasoning
Subcategory: {subcategory_name}
{subcategory_description}

# Design Inputs
- Concrete focus: {class_level}
- Theme seeds: {theme_words}
- Inspiration only, do not copy: {variant_example}

# Required Qualities
1. Invented but plausible: all organizations, policies, products, people,
   places, datasets, cases, and events must be fictional.
2. Domain realism: use the vocabulary, document conventions, and evidence style
   natural to the chosen subcategory.
3. Reasoning density: include definitions, cases, tables, thresholds,
   exceptions, tensions, and cross-section dependencies.
4. Internal consistency: numeric values, dates, categories, and examples must
   agree across the document.
5. Self-containment: all knowledge required for later tasks must appear in the
   document.
6. Length: target at least {min_chars} characters of substantive body text. Do
   not pad.

# Output Rules
- Write in English.
- Output only the document body, with no meta-commentary and no markdown code
  fence.
- Use clear section headings, tables, and examples when useful.
- Do not mention that the document is fictional or benchmark-generated.

Before final output, silently check that the document is complete, internally
consistent, and rich enough to support multi-hop reasoning.
\end{Prompt}

\paragraph{Rule-system context generation.}
For rule-system tasks, the context prompt emphasizes formal precision,
deterministic state transitions, conflict resolution, and priority hierarchies.

\begin{Prompt}
# Role
You are a benchmark context designer for Context-CoT. Your job is to create a
fully original, self-contained rule system that can support difficult multi-step
reasoning tasks.

# Target Category
CL-Bench category: Rule System Application
Subcategory: {subcategory_name}
{subcategory_description}

# Design Inputs
- Concrete focus: {class_level}
- Theme seeds: {theme_words}
- Inspiration only, do not copy: {subcategory_example}

# Required Qualities
1. Originality: invent all entities, rules, symbols, institutions, and examples.
   Do not adapt real programming languages, games, laws, or public systems with
   superficial renaming.
2. Formal precision: define entities, attributes, operations, state transitions,
   conflict priorities, exceptions, and termination conditions clearly enough
   for deterministic task solving.
3. Reasoning density: include interacting rules, delayed effects,
   cross-references, boundary cases, examples, and at least one precedence
   hierarchy.
4. Self-containment: a downstream model must be able to answer later questions
   using only this document.
5. Length: target at least {min_chars} characters of substantive body text. Do
   not pad with repetition.

# Suggested Structure
1. Overview and scope.
2. Vocabulary, symbols, and entity schema.
3. Core mechanics and state transitions.
4. Conflict resolution and priority order.
5. Worked examples, records, calculations, or pseudo-code.
6. Edge cases and invalid states.

# Output Rules
- Write in English.
- Output only the document body, with no meta-commentary and no markdown code
  fence.
- Use stable section headings and precise terminology.
- Keep examples internally consistent with the rules.
- Do not reveal that the document is generated for a benchmark.

Before final output, silently check for contradictions, undefined terms,
impossible examples, and low-information filler.
\end{Prompt}

\paragraph{Empirical-discovery context generation.}
For empirical discovery and simulation tasks, the context prompt requires
invented observations, logs, measurements, or simulation traces from which a
hidden pattern can be inferred.

\begin{Prompt}
# Role
You are a benchmark context designer for Context-CoT. Your job is to create a
fully invented empirical dossier that teaches a new pattern, mechanism, law, or
simulator behavior inside the context itself.

# Target Category
CL-Bench category: Empirical Discovery & Simulation
Subcategory: {subcategory_name}
{subcategory_description}

# Design Inputs
- Concrete focus: {class_level}
- Theme seeds: {theme_words}
- Inspiration only, do not copy: {variant_example}

# Required Qualities
1. Invented empirical world: all variables, entities, instruments, materials,
   locations, and labels must be fictional or generic. Do not rely on real-world
   scientific laws as the answer.
2. Discoverable structure: include enough observations for a downstream model to
   infer a rule, trend, threshold, causal relationship, transition dynamic, or
   validity boundary from the context.
3. Data richness: include tables, logs, repeated trials, measurements,
   ablations, interventions, or simulation traces.
4. Noise and traps: include irrelevant variables, noisy measurements, edge
   cases, conflicting preliminary notes, or regime shifts, while keeping the
   true pattern internally consistent.
5. Self-containment: all definitions, units, measurement conventions, and
   evidence needed for later tasks must appear in the document.
6. Length: target at least {min_chars} characters of substantive body text. Do
   not pad.

# Output Rules
- Write in English.
- Output only the document body, with no meta-commentary and no markdown code
  fence around the whole document.
- Do not include questions, answers, rubrics, benchmark references, or AI-system
  references.
- Do not reveal the hidden rule as a final summary; let it be inferable from the
  evidence.
\end{Prompt}

\paragraph{Procedural-task context generation.}
For procedural task execution, the prompt generates both the system instruction
and the context as a JSON object. This is because procedural tasks require a
deterministic role specification, an execution protocol, and a messy case file
that must be interpreted jointly.

\begin{Prompt}
[Role]
You are a benchmark context designer for Context-CoT. Create a strict
procedure-execution scenario where a model must follow a newly provided manual,
workflow, or operational protocol instead of relying on generic commonsense.

[Target Domain]
- Domain: {domain}
- Sub-domain: {sub_domain}

[Output Shape]
Return only one valid JSON object with this shape:
{
  "system_instruction": "...",
  "context": "..."
}

[System Instruction Requirements]
The system_instruction field must define the procedure-following role, allowed
evidence, and output contract. Include:
- Required order of operations and how to handle skipped prerequisites.
- Rules for conditional branches, exception paths, warnings, and stop
  conditions.
- How to treat missing materials, ambiguous requests, invalid states, and stale
  versions.
- Exact output schemas, labels, checklist fields, status values, or escalation
  phrases.
- Prohibitions against inventing steps, compressing mandatory checks, or using
  external procedure knowledge.

[Context Requirements]
The context field must be a realistic procedural case file. It should include a
manual or protocol excerpt, transcript or operator notes, materials/status
artifacts, version labels, and enough evidence for deterministic downstream
questions.

[Quality Bar]
- The procedure must contain ordered steps, prerequisites, conditional branches,
  and at least one hard stop or escalation condition.
- Artifacts should use raw tables, logs, checklists, forms, or policy excerpts
  without explanatory hints.
- The scenario must require multi-hop reasoning across procedure text and
  execution records.
- Do not add prose outside the JSON object.

Before final output, silently verify that the JSON parses, that the procedure is
enforceable, and that the case supports strict binary rubrics.
\end{Prompt}

\subsection{System-Instruction Construction}

For non-procedural categories, we separately generate a reusable system
instruction paired with each context. The system instruction specifies the
downstream model's role, evidence constraints, answer format, and fallback
behavior.

\begin{Prompt}
[Role]
You are designing a reusable System Instruction for a model that will answer
questions about the generated context.

[Subcategory]
{subcategory_name}
{subcategory_description}

[Behavioral Dimensions]
- Tone and style: {tone}
- Target audience: {audience}
- Reasoning focus: {focus}
- Structure constraint: {structure}

[Instruction Design Requirements]
- Write the instruction in second person, addressing the downstream model as
  "you".
- Give the model a concrete professional, academic, analyst, or domain-specific
  role; avoid generic "AI assistant" phrasing.
- Require the model to answer only from the provided context and not external
  facts.
- Define one exact fallback response for insufficient or underdetermined
  context.
- Include up to {constraint_count} concise, non-overlapping, verifiable
  constraints.
- Make constraints judgeable later: tone, structure, citation, evidence use,
  fallback behavior, numeric precision, or output format.
- Do not include topic-specific facts, examples, or document content.
- Keep the instruction concise.

[Output]
Return only the System Instruction text. Do not add a title, label, code fence,
bullets about your process, or explanatory preface.
\end{Prompt}

\subsection{Task and Rubric Construction}

Given a generated context and its system instruction, we prompt a task generator
to produce context-dependent questions, reference answers, and fine-grained
binary rubrics. The answer field contains only the final user-facing answer and
does not include hidden reasoning or CoT. Rubrics are designed to be atomic and
judgeable, so that a candidate answer is correct only when all rubrics are
satisfied.

\begin{Prompt}
[Role]
You are the lead task architect for Context-CoT / CL-Bench. Create
context-learning tasks from the supplied System Instruction and Context.

[System Instruction]
{system_section}

[Context]
{context}

[Design Objectives]
- Generate exactly {num_tasks} tasks.
- Each task must require synthesizing distributed information from the Context,
  such as definitions, constraints, cases, tables, observations, procedures,
  exceptions, or priority rules.
- Prefer decisions, classifications, calculations, predictions, state
  transitions, operational artifacts, or formatted advisory outputs.
- Do not ask for summaries, vague opinions, trivia, or direct lookup when
  multi-step reasoning is required.
- The answer field is the final user-facing answer only. Do not include hidden
  reasoning or chain-of-thought.

[Golden Answer Requirements]
- Use only the System Instruction and Context; do not import external facts.
- Obey the System Instruction's role, tone, structure, formatting, citation,
  fallback, and refusal behavior.
- Be concise but complete: state the final outcome and include only the support
  required by the System Instruction.
- If rules, observations, or artifacts conflict, apply the stated priority,
  reliability hierarchy, or validity boundary.

[Rubric Requirements]
Write at least {min_rubrics} rubrics for each task. Every rubric must be:
- Binary and atomic.
- Specific to a visible answer property, not a vague quality judgment.
- Grounded in exact context details, such as named entities, values, thresholds,
  dates, sections, units, states, identifiers, or required output fields.
- Able to catch common wrong answers, unsupported outside knowledge, distractor
  use, and non-compliance with the System Instruction.

Rubrics should cover:
1. Correct final outcome, decision, prediction, or artifact.
2. Correct use of key context facts, rules, observations, or procedure steps.
3. Correct handling of exceptions, conflicts, priorities, noise, invalid states,
   or boundary conditions.
4. Required answer format, tone, schema, citations, precision, or fallback
   phrase.
5. Exclusion of unsupported assumptions, hallucinated evidence, or forbidden
   alternatives.

[Output Contract]
Return only valid JSON. Do not use markdown fences, comments, or prose outside
JSON.
Schema:
{
  "tasks": [
    {
      "question": "...",
      "answer": "...",
      "rubrics": [
        "The response ...",
        "The response ..."
      ]
    }
  ]
}

Before final output, silently verify that the JSON parses, that tasks are
non-duplicative, and that rubrics are concrete enough for strict judging.
\end{Prompt}

\subsection{Minimum-Leakage CoT Synthesis}

The minimum-leakage stage generates candidate CoTs without exposing the
reference answer or the full rubric list to the teacher model. The teacher sees
only the system instruction, context, and question. If a candidate fails hidden
verification, only one failed rubric is exposed as minimal feedback in the next
regeneration round. The reference answer remains hidden throughout teacher-side
generation.

\paragraph{Teacher candidate and refinement prompt.}

\begin{Prompt}
Generate one context-grounded reasoning trajectory and final answer.

Rules:
- Use only the provided system instruction, context, and question.
- Do not rely on outside knowledge.
- Do not mention hidden evaluation, rubrics, an oracle, or a reference answer.
- First extract the relevant contextual evidence, then reason over that
  evidence.
- Return strict JSON with exactly two string fields: "think" and "answer".

[System Instruction]
{system_text}

[Context]
{context_text}

[Question]
{question_text}

[Optional Minimal Feedback]
Minimal failed-rubric feedback from the previous attempt. Use only these
criteria as targeted correction signals; do not mention them in your answer:
{one_failed_rubric_if_any}
\end{Prompt}

\paragraph{Hidden rubric judge prompt.}
A separate judge model receives the hidden reference answer and the full rubric
set. It is used only for filtering, not for teacher-side CoT generation.

\begin{Prompt}
System:
You are a strict rubric-based judge for context-learning CoT synthesis.

User:
Determine whether the candidate answer is correct and whether the reasoning is
faithful to the context. Use the hidden reference answer and all rubrics only for
judging.

Return strict JSON:
{
  "passed": true or false,
  "failed_rubric_indices": [1-based rubric indices that failed],
  "rationale": "brief reason"
}

[System Instruction]
{system_text}

[Context]
{context_text}

[Question]
{question_text}

[Hidden Reference Answer]
{reference_answer}

[Full Hidden Rubrics]
{rubric_block}

[Candidate Reasoning]
{candidate_think}

[Candidate Answer]
{candidate_answer}
\end{Prompt}

\subsection{Answer-Exposed CoT Baseline Prompt}

For comparison, we include an answer-exposed CoT distillation baseline. Unlike
Context-CoT, this baseline gives the teacher access to the reference answer
during CoT generation. It therefore tests whether answer-conditioned rationales
are sufficient for improving context learning.

\begin{Prompt}
# Role
You are an expert reasoning demonstration system. Given the context, question,
rubrics, and reference answer, generate a reasoning trajectory and final answer
for supervised fine-tuning.

[Context]
System Content: {system_content}
User Content: {user_content}

[Reference Answer]
{answer}

[Rubrics]
{rubrics}

# Requirements
- The final answer must satisfy the reference answer and rubrics.
- The reasoning should be written as if it is derived from the provided context.
- Do not explicitly mention the reference answer, rubrics, oracle, or hidden
  evaluation in the generated reasoning.
- Use the context as the apparent source of evidence.
- Return the output in the following format:

<think>
{reasoning}
</think>
{final_answer}
\end{Prompt}

\end{document}